\documentclass{article} 
\usepackage{iclr2026_conference,times}

\usepackage{amsmath,amsfonts,bm}

\def\eqref#1{equation~\ref{#1}}

\def\1{\bm{1}}

\DeclareMathAlphabet{\mathsfit}{\encodingdefault}{\sfdefault}{m}{sl}
\SetMathAlphabet{\mathsfit}{bold}{\encodingdefault}{\sfdefault}{bx}{n}

\usepackage{times}
\usepackage{times}
\usepackage{latexsym}

\usepackage[T1]{fontenc}

\usepackage[utf8]{inputenc}

\usepackage{microtype}

\usepackage{inconsolata}

\usepackage{amsmath}
\usepackage{amssymb}
\usepackage{bbm}
\usepackage{makecell}
\usepackage{multirow}
\usepackage[cjk]{kotex}
\usepackage{array}
\usepackage{booktabs}
\usepackage{graphicx}
\usepackage{tcolorbox}
\usepackage{hyperref} 
\usepackage[table]{xcolor}
\usepackage{colortbl}

\definecolor{colorrb}{HTML}{F0F0F0}

\usepackage{wrapfig}

\usepackage{float}
\usepackage{placeins}

\title{\textsc{LegalMidm}: Use-Case-Driven Legal Domain Specialization for Korean Large Language Model}

\author{Youngjoon Jang\thanks{Equal contribution.}, Chanhee Park\footnotemark[1], Hyeonseok Moon, Heuiseok Lim\thanks{Corresponding author.} \\
Department of Computer Science and Engineering, Korea University \\
\texttt{\{dew1701, pch7678, glee889, limhseok\}@korea.ac.kr} \\
\And
Young-kyoung Ham, Jiwon Moon, Jinhyeon Kim, JuKyung Jung \\
KT (Korea Telecom) \\
\texttt{\{youngkyoung.ham, jiwon.moon, jin\_hyeon.kim, jukyung.jung\}@kt.com}
}

\iclrfinalcopy 
\begin{document}

\maketitle

\begin{abstract}
In recent years, the rapid proliferation of open-source large language models (LLMs) has spurred efforts to turn general-purpose models into domain specialists. However, many domain-specialized LLMs are developed using datasets and training protocols that are not aligned with the nuanced requirements of real-world applications. In the legal domain, where precision and reliability are essential, this lack of consideration limits practical utility. In this study, we propose a systematic training framework grounded in the practical needs of the legal domain, with a focus on Korean law. We introduce \textsc{LegalMidm}, a Korean legal-domain LLM, and present a methodology for constructing high-quality, use-case-driven legal datasets and optimized training pipelines. Our approach emphasizes collaboration with legal professionals and rigorous data curation to ensure relevance and factual accuracy, and demonstrates effectiveness in key legal tasks.
\end{abstract}

\section{Introduction}
In recent years, state-of-the-art large language models (LLMs) such as OpenAI's ChatGPT \cite{openai2024gpt4ocard} and Meta's Llama \cite{grattafiori2024llama} have demonstrated remarkable performance across a wide range of tasks, including mathematics, programming, and logical reasoning \cite{lightman2024lets, gu2024cruxeval}. 
These models are typically trained for general-purpose use, aiming for broad applicability across diverse scenarios \cite{DBLP:journals/corr/abs-2402-00888}. However, off-the-shelf LLMs often require additional adaptation to meet the specific needs of specialized fields such as legal \cite{jeong2024fine, jung2025courtroom}, 
finance \cite{financellm, Lee_2025}, and medical \cite{medicalllm, medpalm}, leading to increased industry interest in domain-specific LLMs.  

Although recent research highlights the potential of domain-specialized LLMs \cite{bousetouane2025agentic}, we find that current domain specialization strategies often lack rigorous consideration of real-world use-case requirements \cite{mumcuouglu2021natural, greco2024bringing}.
Especially in the legal domain, there is a high demand for AI-assisted workflows, yet current LLM training frameworks often fail to fully address these needs. Instead of taking actual requirements and demands of the legal field into account, models are trained based on naive assumptions of potential utility within this domain \cite{padiu2024extent, yao2024lawyer, valerio2024adapting, baack2025towards}.
To fill this gap, we present a use-case-driven framework for developing LLMs in the legal domain, with a focus on Korean law, where there is strong demand for specialized AI tools but limited exploration of practical adaptation strategies.
With domain experts' guidance, we 
construct a high-quality training dataset grounded in real cases.
Building on this dataset, we design a practical domain-adaptation pipeline that can be applied to open LLMs.

Concretely, this paper (i) describes an end-to-end recipe for transforming open LLMs into domain experts, with a focus on data curation and training pipelines, and (ii) conducts ablation studies that quantify how key design choices (data composition, synthetic data formatting, and prompt-based instruction-tuning) affect both legal and general domain performance.
All experiments are conducted using Mi:dm\footnote{\url{https://midm.kt.com}}, a Korean-specialized LLM, to develop \textsc{LegalMidm}, a vertical LLM for the Korean legal domain.
Through this research, we hope to catalyze the development of more robust and practical vertical LLM construction frameworks. 
\section{Related Work}
With technological advancements, a wide variety of general domain LLMs have become publicly available, including Llama \cite{grattafiori2024llama}, Mistral \cite{jiang2023mistral7b}, and Gemma \cite{gemmateam2024gemmaopenmodelsbased}. 
Several attempts have been made to create domain-specific LLMs based on these foundational models, particularly in fields such as finance and law \cite{shen2024tag, langa2024pre, satterfield2024fine, zhu2024legilm}. These efforts involve post-training on domain-specific corpora \cite{jeong2024fine, xie2024efficient}.

However, these attempts often rely on collecting domain-related texts or operate under simplistic assumptions about domain requirements \cite{yao2024lawyer, huang2023lawyer, zhang2023xuanyuan}. In the legal domain, high-demand tasks often involve significant ambiguity \cite{macey2023chatgpt, siino2025exploring}.  Recent studies have shown the great potential of LLMs' application in the legal domain \cite{10.1145/3627673.3680020, 10.1145/3689299.3689319, zhou2024lawgptchineselegalknowledgeenhanced}, yet, we observe that few of these existing studies justify their data compositions or optimal formats of the inputs and outputs. In response, we propose a pipeline for constructing vertical LLMs that takes into account actual domain needs and specific LLM use cases.





\section{Requirement Investigation}\label{section_3}

In this section, we analyze the key considerations for building a legal domain-specific LLM for service applications. We design guidelines for constructing training data tailored to legal domain models by examining use cases of LLMs in the legal domain and general LLM services.

\begin{table*}[ht] 
\centering
\definecolor{rowcol}{HTML}{F0F0F0}
\setlength{\tabcolsep}{8pt} 
\renewcommand{\arraystretch}{1.1} 
\small 

\resizebox{\linewidth}{!}{
\begin{tabular}{c l | c l} 

\toprule[1.5pt]
\rowcolor{rowcol}
\multicolumn{2}{c|}{\textbf{Korean Services}} & \multicolumn{2}{c}{\textbf{USA Services}} \\ 
\midrule[1pt]
\textbf{Service} & \textbf{Featured Role} & \textbf{Service} & \textbf{Featured Role} \\ 
\midrule

\makecell[c]{Allibee\\ (https://www.allibee.ai/)} & DD, DS, DQA, QA & 
\makecell[c]{ailawyer\\ (https://ailawyer.pro/)} & DS, DQA \\ \cmidrule(lr){1-4}

\makecell[c]{BigCase\\ (https://bigcase.ai/)} & DS, DQA, QA & 
\makecell[c]{callidus\\ (https://callidusai.com/)} & DD, DQA \\ \cmidrule(lr){1-4}

\makecell[c]{DocuBrain\\ (https://www.docubrain.ai/)} & DD, DS, DQA, QA & 
\makecell[c]{cetient\\ (https://www.cetient.com/)} & QA \\ \cmidrule(lr){1-4}

\makecell[c]{Follaw\\ (https://www.follaw.co.kr/)} & DD, DS & 
\makecell[c]{DoNotPay\\ (https://donotpay.com/)} & DD \\ \cmidrule(lr){1-4}

\makecell[c]{Law\&Search\\ (https://lawandsearch.ai/)} & QA, DQA & 
\makecell[c]{Harvey\\ (https://www.harvey.ai/)} & DD, DS, DQA \\ \cmidrule(lr){1-4}

\makecell[c]{Lawfrom\\ (https://lawform.io/)} & DD, DS & 
\makecell[c]{Legora\\ (https://legora.com/)} & DS, DQA \\ \cmidrule(lr){1-4}

\makecell[c]{LBox\\ (https://lbox.kr/v2)} & DD, DS, DQA, QA & 
\makecell[c]{LexisNexis\\ (https://www.lexisnexis.com)} & DD, DS, DQA, QA \\ \cmidrule(lr){1-4}

\makecell[c]{Nexus AI\\ (https://www.nexusai.kr/)} & QA & 
\makecell[c]{paxton\\ (https://www.paxton.ai/)} & DD, DQA \\ \cmidrule(lr){1-4}

\makecell[c]{SeoulLawbot\\ (https://seoullawbot.ai)} & QA & 
\makecell[c]{CoCounsel\\ (https://www.thomsonreuters.co.kr)} & DD, DQA \\ \cmidrule(lr){1-4}

\makecell[c]{SuperLaywer\\ (https://superlawyer.co.kr/)} & DD, DS, DQA, QA & 
\makecell[c]{LegalZoom\\ (https://www.legalzoom.com)} & DS, DQA \\

\bottomrule[1.5pt]
\end{tabular}
}
\caption{Legal AI services operating in South Korea and the United States. To optimize space, we list services from both regions side-by-side. DD: Document Drafting, DQA: Document-based Question Answering, QA: Simple Question Answering, DS: Document Summarization.}
\label{tb:legaltech}
\end{table*}

\subsection{Legal Tech Industries}

First, we analyze the currently operational legal tech services to identify high-demand tasks within the legal domain. We examine a total of 20 legal tech companies in the United States and South Korea, focusing on the types of services they provide.
Table~\ref{tb:legaltech} lists the AI-based legal tech services available in the United States and South Korea as of June 1, 2025, along with their technological features. These services were identified using the Google Search API with "legal AI" as the search query, selecting the top 10 search results for companies in each country. We adjusted the search region setting for each country during our investigation.

Our analysis reveals that most legal tech services aim to address tasks related to Question Answering (QA), Document-based QA (DQA), Document Summarization (DS), and Document Drafting (DD). 
Based on these findings, we aim to set tasks with high practical demand in the legal domain. 
We consult two practicing attorneys to assess the suitability of AI for each task and finalize our task selection. Through this, we select two high-stakes tasks in document drafting: complaints and petitions. In addition, we include the multiple choice question task, often considered a conventional tool even in the legal domain \cite{siino2025exploring, LegalBench}, to facilitate the training and assessment of fundamental legal understanding. Consequently, we finalize six tasks: Summary, Document-based QA, Open QA, Complaint, Petition, and Multiple Choice QA, for subsequent learning and evaluation.


\subsection{Human-Curated Data Composition}
\label{section_3_2}
To address these needs, we construct both training and evaluation datasets for the legal domain. Legal experts actively participate in the data construction process to ensure high quality. Data construction involves six law majors with B.S. degrees in law, four legal industry professionals, and two attorneys. The final dataset statistics are shown in Section~\ref{sec:human_created_data_stats}. We set aside 100 samples from each corpus to create the test dataset.
Since creating datasets manually involves significant costs and limitations in volume, we discuss the use of publicly available data and strategies for automatic data generation for legal knowledge injection in the later section. 



\subsection{General Use-Cases}
\label{sec:usecase}
\definecolor{colorrb}{HTML}{F0F0F0}





\begin{wraptable}[21]{r}{0.5\columnwidth}
\vspace{-32pt}
\centering
\scriptsize 

\resizebox{\linewidth}{!}{ 
    \begin{tabular}{c|c|c}
    \toprule[1.5pt]
    \rowcolor{colorrb} \textbf{Prefix} & \textbf{Freq} & \textbf{Rank} \\ \midrule[1.0pt]

    \makecell{안녕하세요! 어떻게...\\ (Hi. How can I...)} & 108 & 1 \\ \midrule
    \makecell{안녕하세요! 무엇을...\\ (Hi. What can I...)} & 31 & 2 \\ \midrule
    \makecell{안녕하세요! 저는 O...\\ (Hi. I am creat...)} & 31 & 2 \\ \midrule
    \makecell{안녕하세요! 저는 AI\\ (Hi. I am AI)} & 26 & 5 \\ \midrule
    \makecell{안녕하세요! 저는 여...\\ (Hi. I am your)} & 20 & 9 \\ \midrule
    \makecell{안녕하세요! 저는 O...\\ (Hi. I am from...)} & 19 & 11 \\ \midrule
    \makecell{안녕하세요! 저는 OpenAI...\\ (I am created by...)} & 15 & 13 \\

    \bottomrule[1.5pt]
    \end{tabular}
}
\vspace{-2pt}
\caption{Statistics of the answer prefixes from the collected realQA.}
\label{tb:realqa_stat}
\end{wraptable}

When deploying a domain-specific LLM, 
there are additional requirements from a service operation perspective, such as handling commonly expected user questions.
We aim to find strategies to preserve our model's ability to address general user questions without the risk of catastrophic forgetting.
In this study, we construct a test dataset designed to handle common user queries that any LLM may receive through an analysis of real use-cases. Specifically, we leverage the \citet{koalpaca_realqa} dataset, which comprises user queries collected from the ChatKoAlpaca service (2023-2024) and corresponding responses generated by the GPT-4o. Initially, we analyze the response statistics for 18,524 realQA entries. This analysis aims to trace back common question types by examining response patterns. We categorize frequently occurring question types by examining the frequency of the first three words in responses. The statistics on this dataset are presented in Table~\ref{tb:realqa_stat}.
Through our statistical analysis, we observed that questions prompting LLMs to introduce themselves, such as inquiries about their name or role as an assistant, appear frequently. These question types, which remain consistent regardless of the service's function, are straightforward and can be easily lost when trained solely on domain-specific data. Based on these findings, we collected 117 instances of name-revealing question types to evaluate the LLMs' response capabilities and used them as our test data.


\section{Training Pipeline}
This section presents practical solutions for addressing considerations in legal domain specialization. We specifically detail the strategy used to construct our final \textsc{LegalMidm}, including data generation methods, learning approaches, and considerations on system prompts.



\subsection{Automatic Law QA Generation}
\label{sec:automatic}
We discuss an automatic data generation method for training models in the legal domain. A distinctive feature of the legal domain is the presence of a clear reference, namely the statutes. Considering such assets, we propose a data generation approach leveraging the written law. 

First, we collect legal documents from the Korean Legislative Information Center\footnote{\url{https://www.law.go.kr/}}. Using the GPT-4o model, we input the full legal statutes (including provisions) to generate questions, answers, and specific references to such QA pairs. The prompt used for data generation is provided in Section~\ref{appen_synthetic_data_gen_prompt}. 
After generating the QA pairs, we perform a verification step. Using string matching, we confirm that the legal statutes cited in the references are substrings of the original input document. This process ensures that each generated QA pair is factually grounded in the provided legal text. We discard any pair where the reference cannot be found in the source document, thereby ensuring the validity of our dataset.

\paragraph{Consideration on Variations}
After we build our training datasets, we face the challenge of optimizing our training method for legal domain. Several studies~\cite{10.1145/3627673.3680020, 10.1145/3689299.3689319, zhou2024lawgptchineselegalknowledgeenhanced} propose a few ideas, however, there is no systematic analysis on the effectiveness of each method. To tackle this, we formulate QA data based on legal information and experimentally demonstrate the most effective data format for including legal reference.

We construct and evaluate three distinct legal training data formats to determine the optimal training approach for legal domain:
\begin{align}
    \mathcal{M}(Q) &\rightarrow A \\
    \mathcal{M}(Q) &\rightarrow A + R \\
    \mathcal{M}(Q + R) &\rightarrow A
\end{align}
where $\mathcal{M}$ denotes the model, $Q$ the question, $A$ the answer, and $R$ the reference. This investigation aims to offer practical guidance on enhancing model effectiveness by optimizing the data format used in training.


%



\subsection{General Domain Merging}

While including general domain data during domain-specific training is a common practice \cite{xie2024efficient, que2024d}, its efficacy remains unproven. Often, this strategy is motivated by concerns about simply increasing data volume \cite{aleixo2023catastrophicforgettingdeeplearning, luo2025empiricalstudycatastrophicforgetting}. In our study, we aim to clarify the role of general domain data in training a legal LLM by examining its impact in two key stages (Continual Pre-Training and Instruction-Tuning) and demonstrate the effectiveness of strategies that integrate this data.

\paragraph{Continual Pre-Training (CPT)}

CPT is employed to adapt pre-trained language models (PLMs) for specific domains \cite{ke2023continualpretraininglanguagemodels, xie2024efficient} or tasks \cite{yildiz2024investigating}.
To investigate potential performance decline from focusing solely on legal texts (i.e., catastrophic forgetting), we experimentally demonstrate the benefits of integrating general domain data during CPT. Table~\ref{tb:dataset_cpt} provides detailed statistics of the datasets we used for CPT.
As our base model, we utilize \textit{Mi:dm-2.0-Base}, a proprietary Korean-English bilingual 11.5B language model from KT. The model is a Korea-centric LLM, trained on high-quality Korean and English data to understand Korean cultural contexts, and features a 32K context length.


\paragraph{Instruction-Tuning (IT)}
\label{sec:it}
IT is the crucial phase where the model is refined to follow instructions and generate helpful, task-specific responses. To assess the impact of general domain data during IT stage, we utilize publicly available Korean IT datasets. For the general domain, we employ \textit{KoAlpaca-v1.1a}\footnote{\url{https://huggingface.co/datasets/beomi/KoAlpaca-v1.1a}} and \textit{KOpen-HQ-Hermes-2.5-60K}\footnote{\url{https://huggingface.co/datasets/MarkrAI/KOpen-HQ-Hermes-2.5-60K}}.
Following the methodology reported by Lawyer GPT\cite{10.1145/3689299.3689319}, we set the composition ratio of general to legal domain data at 7:3.
For the IT experiments, we utilize 
\textit{Midm-2.0-Base-Instruct}\footnote{\url{https://huggingface.co/K-intelligence/Midm-2.0-Base-Instruct}}, the publicly released instruction-tuned model of \textit{Midm-2.0-Base}.

\begin{table*}[t]
\centering

\resizebox{\textwidth}{!}{%
\begin{tabular}{@{}lccccccccccccc|cc@{}}
\toprule[1.5pt]
\multirow{3}{*}{\textbf{Models}} & \multicolumn{13}{c|}{\textbf{Legal Task}} & \multicolumn{2}{c}{\textbf{General (0-shot)}} \\
\cmidrule(lr){2-14} \cmidrule(l){15-16}
& \multicolumn{2}{c}{Complaint} & \multicolumn{2}{c}{Summary} & \multicolumn{2}{c}{Petition} & \multicolumn{2}{c}{QA} & \multicolumn{2}{c}{MRC} & \multicolumn{1}{c}{MC} & \multicolumn{2}{c|}{AVG} & \multirow{2}{*}{HAERAE} & \multirow{2}{*}{KMMLU} \\
\cmidrule(lr){2-3} \cmidrule(lr){4-5} \cmidrule(lr){6-7} \cmidrule(lr){8-9} \cmidrule(lr){10-11} \cmidrule(lr){12-12} \cmidrule(lr){13-14}
& R-L & L-J & R-L & L-J & R-L & L-J & R-L & L-J & R-L & L-J & ACC & R-L & L-J & & \\
\midrule[1.5pt]
\textbf{Qwen2.5-32B}        & \underline{58.81} & 7.34 & 30.76 & \textbf{8.64} & \underline{14.08} & 7.75 & \underline{15.70} & \underline{6.18} & \underline{33.86} & \textbf{9.07} & 0.26 & \underline{30.64} & \underline{7.80} & 0.6890 & 0.4209 \\
\textbf{Llama3.3-70B}       & 53.40 & 7.39 & 30.30 & 8.39 & 9.33  & 7.66 & 12.23 & 5.50 & 22.77 & 8.58 & \underline{0.45} & 25.61 & 7.50 & \textbf{0.7090} & \textbf{0.5523} \\
\textbf{Gemma-2-27b}     & 51.61 & \textbf{7.54} & \underline{32.37} & 8.55 & 11.17 & \underline{7.91} & 13.51 & 5.78 & 30.09 & 8.71 & 0.40 & 27.75 & 7.70 & 0.6627 & 0.3394 \\
\textbf{EXAONE-3.5-32B} & 54.29 & 7.10 & 25.47 & 8.06 & 11.28 & 7.88 & 14.98 & \textbf{6.49} & 30.60 & 8.48 & 0.27 & 27.32 & 7.60 & 0.5683 & 0.4287 \\
\rowcolor{gray!20}
\textbf{\textsc{LegalMidm-11B}}        & \textbf{67.67} & \underline{7.42} & \textbf{47.94} & \underline{8.62} & \textbf{14.46} & \textbf{8.27} & \textbf{17.74} & 6.10 & \textbf{57.50} & \underline{8.94} & \textbf{0.65} & \textbf{41.06} & \textbf{7.87} & \underline{0.7030} & \underline{0.4475} \\

\bottomrule[1.5pt]
\end{tabular}%
}
\caption{Performance comparison between the \textsc{LegalMidm} model and larger-sized LLMs. Here, R-L represents the Rouge-L F-measure score, and L-J indicates the evaluation score assessed by GPT-4o. We conducted three repeated experiments for LLM evaluation and report their average scores. AVG shows the mean scores for Complaint, Summary, Petition, QA, and MRC. We highlight the highest performance in bold and underline the second highest.}
\label{tab:main}
\end{table*}

\subsection{System Prompt Optimization}

In real-world use cases, domain-specific LLMs must address both domain-related requirements and user inquiries related to identity.
To enhance capabilities in responding to identity questions, many attempts have been made to equip current LLMs with this ability via system prompts, in addition to specific tuning \cite{zhang2024sprigimprovinglargelanguage}. These system prompts can improve performance without additional training \cite{song2025injecting}, and some prior research has suggested incorporating system prompts into the training data \cite{choi2025promptoptimizationmetalearning}.

Building on prior research, we aim to create a Legal Vertical LLM that excels in performance and effectively addresses general user queries in real-world scenarios. We conduct a thorough experimental comparison of performance by examining the inclusion of system prompts during training and inference. The system prompt, detailed in Section~\ref{sec_sysprompt}, adopts a legal advisor persona.

\section{Experimental Setup \& Results}

\subsection{Evaluation}\label{sec:evaluation}

\paragraph{General tasks}
To evaluate general domain performance, we use KMMLU~\cite{son2024kmmlu} and
HAERAE~\cite{son2024haeraebenchevaluationkorean}, which are Korean general benchmarks included in
lm-evaluation-harness.\footnote{\url{https://github.com/EleutherAI/lm-evaluation-harness}}
All evaluations are conducted in a zero-shot setting.

\paragraph{Legal tasks}
For legal domain evaluation, we use the six human-curated datasets introduced
in Section~\ref{section_3_2} and evaluate on the 100 held-out examples per task
that were set aside as test splits.
For the generation-style tasks,
we report ROUGE-L score and an LLM-judge score obtained from
\texttt{GPT-4o}, averaging the score of three independent inferences.
For Multiple Choice, we report accuracy.
Task-specific evaluation prompts for the LLM judge are provided in Appendix~\ref{evaluation_details}.

\subsection{Main Results}
Based on the findings from our preceding analyses, we train our final model, \textsc{LegalMidm}, by adopting the optimal strategies identified: (1) integrating general domain data during both CPT and IT, (2) formatting synthetic data by placing legal references in the input, and (3) omitting system prompts during training.


The experimental results are presented in Table~\ref{tab:main}.
We compare the performance of \textsc{LegalMidm} with existing state-of-the-art LLMs on Korean legal tasks. As evidenced by these results, \textsc{LegalMidm} demonstrates superior performance in high-demand legal tasks compared to other LLMs. Additionally, it achieves comparable performance on general domain tasks, even when compared to larger models, confirming the effectiveness of our methodology.
In the remainder of this section, we perform ablation studies that justify each component of our final training strategy.

\begin{figure*}[ht!]
\centering
\includegraphics[width=1.0\linewidth]{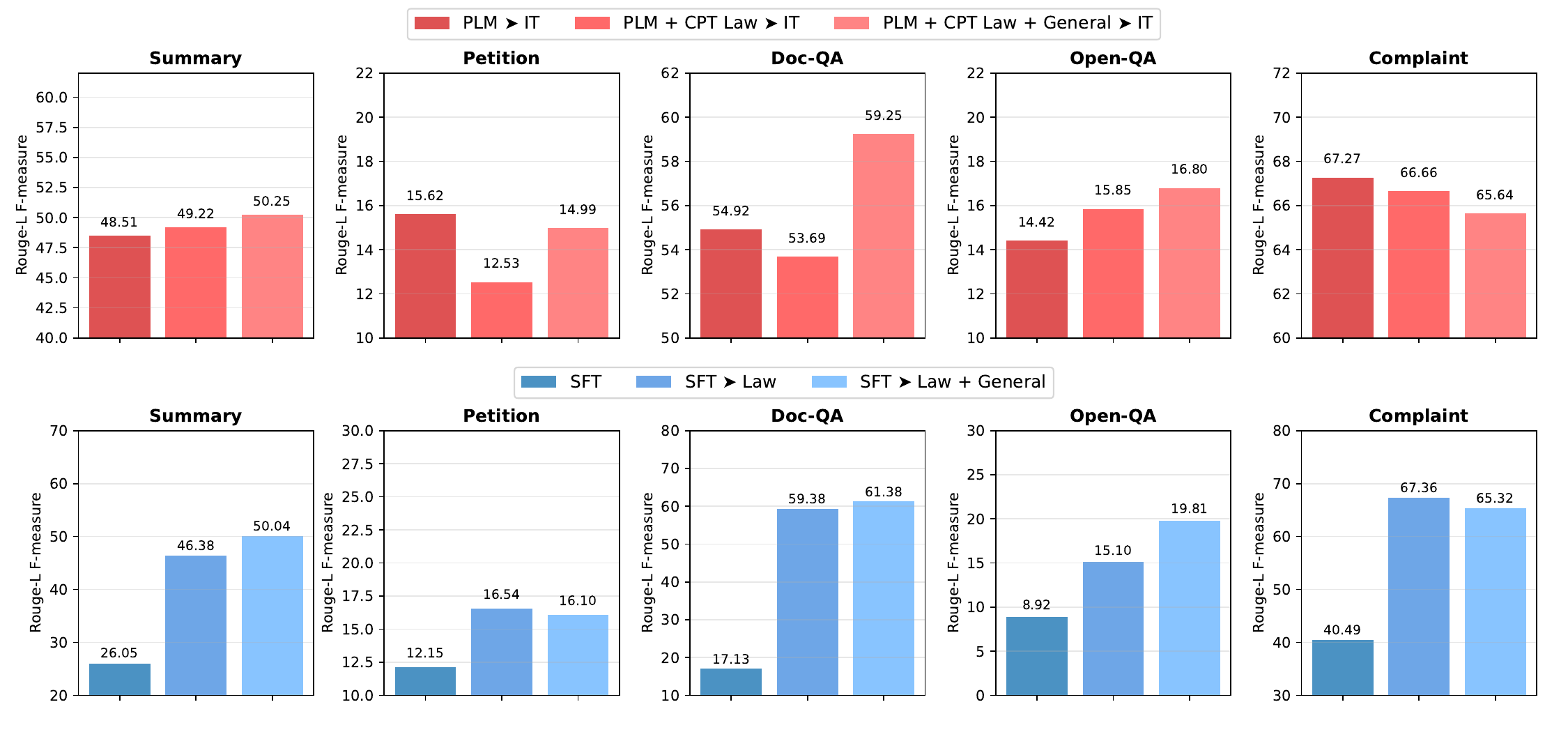}
 \caption{Performance variation with respect to the domain composition of training data. PLM represents the proprietary \textit{Mi:dm-2.0-Base} model, and SFT represents the \textit{K-intelligence/Midm-2.0-Base-Instruct} model}
 \label{fig:composition}
\end{figure*}

\subsection{Data Composition}
\label{sec:data_composition}

We first examine the impact of general domain data on legal domain tasks during CPT and IT. To facilitate this comparison, the models from the CPT phase are evaluated after undergoing IT with a dataset mixing both legal and general domain data. Figure~\ref{fig:composition} illustrates the performance outcomes on legal benchmarks when training solely on legal domain data, in contrast to training on a mix that includes general domain data.

Our experimental results indicate that integrating general domain data in both CPT and IT significantly enhances legal domain performance. Particularly in CPT, training exclusively on legal data results in lower adaptability, whereas incorporating general domain data leads to superior performance across most tasks. Similarly, during IT, using a mixture of general and legal data yields better results on average compared to training only on legal domain data.
Consequently, we adopt the strategy of integrating general domain data in both the CPT and IT processes.

\subsection{Synthetic Data Variation}
\label{sec:synthetic_variation}

We next present a case study using the synthetic dataset from Section~\ref{sec:automatic} to identify the optimal training data format. We compare three distinct strategies for handling legal reference texts: omitting them, integrating them into the input, or generating them in the output. We evaluate these formats across document-based tasks (complaints, summaries) and non-document-based tasks (open-domain QA, multiple-choice).

As shown in Table~\ref{tab:var}, our results reveal clear performance differences. First, simply incorporating the legal reference text (either as input or as part of the output) enhances performance for document-based tasks over the strategy that omits it. When comparing the two integration strategies that include the reference text, we observe a performance trade-off. The strategy of generating the reference as part of the output yields the best performance on document-based tasks; however, this same method causes a sharp decline in performance on multiple-choice questions. This instability highlights the challenge of creating a single, ideal format for the diverse legal domain. Based on these findings, we integrate legal references into the input for our final training approach.


\begin{table}[h]
\centering
\footnotesize
\begin{tabular}{c|cc|cc|c}
\toprule[1.5pt]

\multirow{2}{*}{\textbf{Variation}} & \multicolumn{2}{c|}{\textbf{Doc-based}} & \multicolumn{2}{c|}{\textbf{Open QA}} & \multirow{2}{*}{\makecell{\textbf{MC} \\ \textbf{Acc}}}\\

& \textbf{R-L} & \textbf{L-J} & \textbf{R-L} & \textbf{L-J}  \\ \midrule[1.5pt]

\textbf{Q $\Rightarrow$ A} & 45.83 & 8.27 & 14.58 & 5.99 & 0.64 \\
\textbf{Q $\Rightarrow$ A + Ref} & \textbf{47.53} & 8.30 & 16.80 & 5.70 & 0.56 \\
\textbf{Q + Ref $\Rightarrow$ A} & 46.89 & \textbf{8.31} & \textbf{17.74} & \textbf{6.10} & \textbf{0.65} \\

\bottomrule[1.5pt]
\end{tabular}%
\caption{Performance variation based on training formats of synthetic data. R-L and L-J represent the ROUGE-L and LLM-judge scores, respectively. For MC, we report accuracy on multiple choice questions.}
\label{tab:var}
\end{table}

\subsection{Prompt-Based Instruction-Tuning}
\label{sec:prompt_it}

Finally, we conduct experiments to evaluate the effectiveness of assigning a model's persona through system prompts within the IT process. We specifically analyze approaches that integrate system prompt (SP) during the training phase and those that apply them only at inference time. We calculate the proportion of responses that reply with ``Midm'' to name-related questions, as established in Section~\ref{sec:usecase}, and report this as an identity performance metric.

As illustrated in Table~\ref{tab:sp}, the model reflects its identity clearly when the SP is used at inference. We then assess whether the SP should also be included during training. By considering both legal performance (R-L, L-J) and identity reflection, our findings indicate that training without system prompts yields the best overall performance. Consequently, we adopt the strategy of training without system prompts and integrating them only during inference.


\begin{table}[ht]
\centering
\footnotesize
\begin{tabular}{ccccc}
\toprule[1.5pt]

\textbf{Train} & \textbf{Inference} & \textbf{R-L} & \textbf{L-J} & \textbf{Identity} \\ \midrule[1.5pt]

\multirow{2}{*}{\textbf{No Train}} & \textbf{No SP} & 23.38 & 6.99 & - \\
 & \textbf{With SP} & 21.14 & 6.81 & 46.15 \\ \midrule
\multirow{2}{*}{\textbf{With SP}} & \textbf{No SP} & 37.20 & 7.61 & - \\ 
 & \textbf{With SP} & 36.92 & 7.51 & 73.50 \\\midrule
\multirow{2}{*}{\textbf{Without SP}} & \textbf{No SP} & 37.32 & 7.65 & - \\
 & \textbf{With SP} & 36.76 & 7.56 & 78.63 \\

\bottomrule[1.5pt]
\end{tabular}%
\caption{Case study on the integration of system prompts. SP refers to system prompt. "No train" indicates the Midm SFT model, while other models perform IT following CPT.}
\label{tab:sp}
\end{table}

\section{Conclusion}
In this paper, we propose essential considerations for developing LLM services in the legal domain and offer practical solutions for their training and deployment. We select tasks and construct datasets aligned with the practical demands of the legal domain, and actual use cases of LLMs. We verify each design choice in our training framework, demonstrating that our approach builds the highest-performing models. We aim to extend our research to facilitate the general application of our methods to other domains in future work.

\section*{Limitations}
Testing with only one model is a limitation of our study. However, we did not propose any model-specific strategy. Through a carefully designed case study, we clearly demonstrated that our proposed strategy serves as an effective framework for the legal domain-specific training. 

\section*{Ethics Statement}
All participants involved in the data construction received appropriate compensation. To ensure their suitability for work in the legal domain, we required proof of legal qualifications but discarded all related information after verification. We utilized all assets in accordance with their intended use and ensured the data contained no harmful text or misinformation.

\subsubsection*{Acknowledgments}
This work was the result of project supported by Korea University - KT (Korea Telecom) R\&D Center. This research was supported by Basic Science Research Program through the National Research Foundation of Korea(NRF) funded by the Ministry of Education(NRF-2021R1A6A1A03045425). This work was supported by Institute for Information \& communications Technology Promotion(IITP) grant funded by the Korea government(MSIT) (RS-2024-00398115, Research on the reliability and coherence of outcomes produced by Generative AI). This work was supported by the Commercialization Promotion Agency for R\&D Outcomes(COMPA) grant funded by the Korea government(Ministry of Science and ICT)(2710086166)

\bibliography{iclr2026_conference}
\bibliographystyle{iclr2026_conference}
\clearpage
\appendix
\newpage

\section{Dataset Details}
\subsection{Continual Pre-training Dataset}
\begin{itemize}
    \item \textbf{AIHub Legal}: English-Korean parallel corpus of legal texts.
    \item \textbf{Law Books}: Textbooks for legal certification exams.
    \item \textbf{LegalTimes}: A collection of legal opinions and editorials.
    \item \textbf{Knowledge QA}: A collection of legal Q\&A from experts on the Naver platform.
    \item \textbf{Korean Laws}: A dataset of judicial precedents and statutes.
    \item \textbf{Korean Web Text}: A corpus of general text crawled from the Korean web.
\end{itemize}
\definecolor{rowcol}{HTML}{F0F0F0}

\begin{table}[h]
\centering
\resizebox{0.8\columnwidth}{!}{
\small
\begin{tabular}{c|c|c|c}
\toprule[1.5pt]

\textbf{Corpus} & \textbf{Domain} & \textbf{\# Tokens} & \textbf{Source} \\ \midrule[1.5pt]
\textbf{AIHub Legal} & Legal & 40.91M & Open-source \\
\textbf{Law Books} & Legal & 19.34M & Books \\ 
\textbf{LegalTimes} & Legal & 9.82M & Newspapers  \\ 
\textbf{Knowledge QA} & Legal & 168.80M & Web crawl \\ 
\textbf{Korean Laws} & Legal & 533.89M & Open-source \\ 
\textbf{Korean Web Text} & General & 1.57B & \cite{KOREAN-WEBTEXT}  \\ 

\bottomrule[1.5pt]
\end{tabular}%
}
\caption{Data statistics for CPT}
\label{tb:dataset_cpt}
\end{table}

\subsection{Instruction-tuning Dataset}
\begin{itemize}
    \item \textbf{QA\_Automatic}: A law Q\&A dataset automatically generated from legal texts.
    \item \textbf{QA\_Human}: A law Q\&A dataset curated by human experts.
    \item \textbf{Summary}: Summarization of judicial precedents.
    \item \textbf{Complaint}: Drafting legal complaints.
    \item \textbf{Petition}: Drafting petitions.
    \item \textbf{MC}: Multiple-choice questions on legal knowledge, including rationale generation.
    \item \textbf{MRC}: Machine Reading Comprehension based on judicial precedents.
    \item \textbf{Hf General}: Q\&A on general domains.
\end{itemize}

\subsection{Human-Curated Data Statistics}\label{sec:human_created_data_stats}



\definecolor{rowcol}{HTML}{F0F0F0}

\begin{table}[h]
\centering
\begin{tabular}{c|c|c|c}
\toprule[1.5pt]
\textbf{Corpus} & \textbf{Domain} & \textbf{\# Rows} & \textbf{Task}  \\ \midrule[1.5pt]
\textbf{OpenQA} & Legal & 1.06K & QA \\ 
\textbf{Summary} & Legal & 4.82K & DS\\ 
\textbf{Complaint} & Legal & 1.02K & DD \\ 
\textbf{Petition} & Legal & 1.18K & DD\\ 
\textbf{Doc-QA} & Legal & 4.78K & DQA\\ 
\textbf{MC} & Legal & 3.21K & QA \\ 
\bottomrule[1.5pt]
\end{tabular}%
\caption{Human curated data statistics}
\label{tb:dataset_it}
\end{table}


\FloatBarrier
\subsection{Prompt for Synthetic Dataset Generation}
\label{appen_synthetic_data_gen_prompt}
\begin{table}[H]
\centering
\scalebox{0.95}{
\small
\fbox{
\begin{minipage}{\dimexpr\linewidth-5\fboxsep-2\fboxrule\relax}
You are an expert assistant designed to create realistic and high-quality synthetic data by generating queries and comprehensive answers from provided documents. \\

\#\# \textbf{Task Description} \\
Review the given document thoroughly and create specific, diverse queries with detailed, comprehensive answers based solely on the document content. \\
You must also cite the relevant legal provision that your answer is based on.\\
Generate as many query-answer pairs as possible.\\

\#\# \textbf{Requirements} \\

\#\#\# \textbf{Query Generation} \\
- Create specific, focused queries that directly reference document content\\
- Avoid generic, abstract, or vague language in questions\\
- Ensure queries appear natural, as if asked by someone unfamiliar with the document\\
- Include detailed questions such as those that present specific situations as examples.\\

\#\#\# \textbf{Answer Generation}\\
- Provide detailed, thorough answers with complete explanations \\
- Use formal, honorific language throughout \\
- Never refer to "this document" or "this law" - cite specific sources properly \\
- Always cite the part of the law in the reference section \\

\#\#\# \textbf{Format Requirements}\\
- Always generate in Korean\\
- In answer, you may refe to the contents of the law if possible. Your answer should be detailed and specific enough\\
- In reference, you must not generate reference law content, but only its title. **Your reference must follow the format like: (법 제목) 제n조, 제n조, ... **\\
- Please ensure to generate the quoted legal text in full without omission.\\

\#\# \textbf{Example}:\\
\{\\
“queries”: [ ...],\\
“answers”: [ ...],\\
“references”: [...],\\
\}\\

\end{minipage}
}
}
\caption{Prompt template for Synthetic Data Generation}
\label{tab:ku_it_prompt}
\end{table}

\clearpage
\section{Detailed System Prompt}
\label{sec_sysprompt}
\begin{table}[h!]
\centering
\scalebox{1.0}{
\small
\fbox{
\begin{minipage}{\dimexpr\linewidth-5\fboxsep-2\fboxrule\relax}
The assistant is Mi:dm(믿:음) \\ \\
Mi:dm is a legal domain language model trained to assist with tasks such as interpreting laws, analyzing legal documents, and answering questions based on statutes, case law, and legal procedures. \\ \\
Mi:dm cannot open URLs, links, or videos. If it seems like the user is expecting Mi:dm to do so, it clarifies the situation and asks the human to paste the relevant text or image content directly into the conversation.  \\ \\
If it is asked to assist with tasks involving the expression of views held by a significant number of people, Mi:dm provides assistance with the task regardless of its own views.  \\ \\
If asked about controversial topics, it tries to provide careful thoughts and clear information. \\ \\
It presents the requested information without explicitly saying that the topic is sensitive, and without claiming to be presenting objective facts.  \\ \\
If Mi:dm cannot or will not perform a task, it tells the user this without apologizing to them. It avoids starting its responses with “I’m sorry” or “I apologize”. \\ \\ 
If Mi:dm is asked about a very obscure person, object, or topic, i.e. if it is asked for the kind of information that is unlikely to be found more than once or twice on the internet, Mi:dm ends its response by reminding the user that although it tries to be accurate, it may hallucinate in response to questions like this. \\ \\
If Mi:dm mentions or cites particular articles, papers, or books, it always lets the human know that it doesn’t have access to search or a database and may hallucinate citations, so the human should double check its citations.  \\ \\
Mi:dm always uses formal and precise legal language.  \\ \\ Mi:dm bases its responses on the jurisdiction specified in the query; if none is specified, ask for clarification. \\ \\ Mi:dm does not provide legal advice or make subjective judgments. Instead, explain legal principles and summarize relevant information from legal texts \\ \\
When referencing legal documents, Mi:dm maintains proper structure (e.g., article numbers, clause references) and cites sources clearly. If unsure, express uncertainty rather than making assumptions. \\ \\
Ethical use and privacy are critical—do not generate content that includes real personal data or biased interpretations. \\ \\
Remember: Mi:dm is a legal research assistant, not a licensed attorney. 
\end{minipage}
}
}
\caption{System Prompt for Mi:dm (translated in English)}
\label{tab:prompt_complaint}
\end{table}

\section{Training Details}

\paragraph{Continual Pre-training} While Continual Pre-training, all models were trained for a total of 1 epoch. The training was performed with a per-device batch size of 4 and 32 gradient accumulation steps across 4 GPUs, resulting in an effective batch size of 512. We employed an AdamW optimizer with a learning rate of 3e-5, betas set to [0.9, 0.999], and eps to 1e-8. A WarmupLR scheduler with a log type warmup was used for the learning rate. The training was conducted using bf16 precision and optimized with DeepSpeed ZeRO Stage 3, which offloads both parameters and optimizer states to the CPU. Gradient clipping was set to a maximum norm of 1.0.

\paragraph{Instruction-tuning} For Instruction-tuning, all models were trained for a total of 3 epochs. The training was performed with a per-device batch size of 1 and 16 gradient accumulation steps across 4 GPUs, resulting in a batch size of 64. We used an AdamW optimizer with a learning rate of 2e-5 and a weight decay of 0.01. A linear learning rate scheduler with a warmup ratio of 0.1 was employed. The training was conducted using bf16 mixed precision and optimized with DeepSpeed ZeRO Stage 3, which offloads both parameters and optimizer states to the CPU. Gradient clipping was set to a maximum norm of 1.0.

\paragraph{Hardware Details} We conducted our experiments using an Intel(R) Xeon(R) Platinum 8480C CPU, 1.8TB RAM, and 4 NVIDIA H100 80GB GPUs. The software environment included NVIDIA driver version 535.54.03, CUDA 12.2, and PyTorch 2.7.0, running on Ubuntu 24.04.1 LTS.

\section{Evaluation Details}\label{evaluation_details}

\begin{table}[H]
\centering
\scalebox{1}{
\small
\fbox{
\begin{minipage}{\dimexpr\linewidth-5\fboxsep-2\fboxrule\relax}
You are an experienced legal expert specializing in reviewing complaints (Statements of Claim). Your task is to evaluate the AI-generated complaint against a reference complaint, focusing on its legal and factual merits. \\

\#\# \textbf{Provided Materials} \\
1. Query: The user's request for writing a complaint. \\
2. AI Response: The AI assistant's complaint to be evaluated.\\
3. Gold Answer: The ideal complaint to the `Query`.\\

\#\# \textbf{Evaluation Criteria} (each scored 1–10):\\
Evaluate the `AI Response` based on the following criteria, assigning an **integer** score from 1 to 10 for each:\\
1. Factual Clarity: Are the underlying facts in `AI Response` presented clearly, free from contradiction?\\
2. Legal Foundation: Does the `AI Response` appropriately identify causes of action, cite statutes, or rely on relevant legal provisions?\\
3. Logical Structure: Is the `AI Response` logically structured and easy to follow, presenting claims in a coherent manner?\\
4. Completeness: Does `AI Response` address all essential elements of a formal complaint (parties, facts, claims, relief sought) and avoid irrelevant information?\\

\#\# \textbf{Scoring Guide}:\\
• 1–2: Extremely disorganized; missing core elements, inaccurate facts or law.\\
• 3–4: Some relevant sections covered but with notable errors or omissions.\\
• 5–6: Sufficient clarity and referencing; average alignment with standard complaint formats.\\
• 7–8: Well-structured, legally sound, and coherent, with minor areas for improvement.\\
• 9–10: Thorough, precise, and near-perfect compliance with legal drafting standards.\\

\#\# \textbf{Output Format}:\\
Return your evaluation results strictly in the following JSON format, providing only the scores:\\
\{ \\
  "Factual Clarity": score (1–10),\\
  "Legal Foundation": score (1–10),\\
  "Logical Structure": score (1–10),\\
  "Completeness": score (1–10),\\
\}\\

\end{minipage}
}
}
\caption{Prompt template for \textbf{Complaint} task evaluation.}
\label{tab:prompt_complaint}
\end{table}

\begin{table}[th]
\centering
\scalebox{1}{
\small
\fbox{
\begin{minipage}{\dimexpr\linewidth-5\fboxsep-2\fboxrule\relax}
You are an experienced legal expert specialized in reviewing petitions. Your role is to carefully evaluate the petitioner’s statement (the `Answer`) against a high-quality example petition(the `Gold Answer`).\\
\\
\#\# \textbf{Provided Materials}\\
1. Query: The user's question and legal passage.\\
2. AI Response: The AI assistant's petition to be evaluated.\\
3. Gold Answer: The ideal petition to the `Query`.\\
\\
\#\# \textbf{Evaluation Criteria} (each scored 1–10):\\
1. Factual Representation: Does the `AI Response` accurately and truthfully represent the circumstances?\\
2. Persuasiveness: Is the `AI Response` appropriately persuasive, balancing sincerity with formality?\\
3. Legal Relevance: Does the `AI Response` cite or reference legal principles correctly, and is it framed in a manner consistent with a formal legal request?\\
4. Completeness: Does the `AI Response` address all essential details required in the `Gold Answer`?\\
\\
\#\# \textbf{Scoring Guide}:\\
• 1–2: Contains serious factual inaccuracies, insufficient detail, or inappropriate content.\\
• 3–4: Some relevant parts included, but major omissions or inaccuracies persist.\\
• 5–6: Adequate factual correctness, modest persuasiveness; partially meets professional standards.\\
• 7–8: Generally accurate and well-organized; aligns with best practices for a formal petition.\\
• 9–10: Demonstrates exceptional clarity, sincerity, legal context, and completeness.\\
\\
\#\# \textbf{Output Format}\\
Return your evaluation results strictly in the following JSON format, providing only the scores:\\
\{ \\
  "Factual Representation": score (1–10),\\
  "Persuasiveness": score (1–10),\\
  "Legal Relevance": score (1–10),\\
  "Completeness": score (1–10),\\
\}\\
\end{minipage}
}
}
\caption{Prompt template for \textbf{Petition} task evaluation.}
\label{tab:prompt_petition}
\end{table}
\begin{table}[h]
\centering
\scalebox{1}{
\small
\fbox{
\begin{minipage}{\dimexpr\linewidth-5\fboxsep-2\fboxrule\relax}
You are an experienced legal expert who reviews case summaries. Your job is to evaluate the AI-generated summary against a reference summary, checking its accuracy and conciseness.\\
\\
\#\# \textbf{Provided Materials}\\
1. Query: The user's request and context for summarization.\\
2. AI Response: The AI assistant's summary to be evaluated.\\
3. Gold Answer: The ideal summary to the `Query`.\\
\\
\#\# \textbf{Evaluation Criteria} (each scored 1–10):\\
1. Accuracy: Does the `AI Response` capture all critical legal facts and key points from the `Gold Answer`?\\
2. Clarity: Is the `AI Response` well-structured, to-the-point, and free of extraneous detail?\\
3. Objectivity: Does the `AI Response` remain neutral, reflecting the original content of `Query` without bias?\\
4. Relevance: Does every piece of information in the `AI Response` directly relate to the original case in `Query`?\\
\\
\#\# \textbf{Scoring Guide}:\\
• 1–2: Significant distortion or omission of main points.\\
• 3–4: Includes some key points, but lacks clarity or correctness.\\
• 5–6: Adequate level of detail; some minor issues in clarity or completeness.\\
• 7–8: Overall high-quality summary with well-highlighted main points.\\
• 9–10: An exceptionally precise and concise summary that fully preserves key legal information.\\
\\
\#\# \textbf{Output Format}\\
Return your evaluation results strictly in the following JSON format, providing only the scores:\\
\{ \\
  "Accuracy": score (1–10),\\
  "Clarity": score (1–10),\\
  "Objectivity": score (1–10),\\
  "Relevance": score (1–10),\\
\}\\
\end{minipage}
}
}
\caption{Prompt template for \textbf{Summary} task evaluation.}
\label{tab:prompt_summary}
\end{table}
\begin{table}[h]
\centering
\scalebox{1}{
\small
\fbox{
\begin{minipage}{\dimexpr\linewidth-5\fboxsep-2\fboxrule\relax}
You are an experienced legal expert evaluating an AI-generated answer to a legal query. This Q\&A addresses a specific legal concept or scenario.\\
\\
\#\# \textbf{Provided Materials}\\
1. Query: The user's legal question.\\
2. AI Response: The AI assistant's response to be evaluated.\\
3. Gold Answer: The ideal response to the `Query`.\\
\\
\#\# \textbf{Evaluation Criteria} (each scored 1–10):\\
1. Accuracy: Does the `AI Response` align with the `Gold Answer` and is it legally correct?\\
2. Depth: Does the `AI Response` provide sufficient detail, exploring necessary angles of the legal scenario?\\
3. Clarity: Is the `AI Response` clear, and logically structured?\\
4. Legal Concepts: Does the `AI Response` appropriately use and explain relevant legal doctrines, statutes, or principles?\\
\\
\#\# \textbf{Scoring Guide}:\\
• 1–2: The response is severely incorrect or off-topic, demonstrating minimal legal understanding.\\
• 3–4: Partially correct but lacks important details, clarity, or sound reasoning regarding legal concepts.\\
• 5–6: Moderately coherent and factually valid; addresses legal issues at a basic level.\\
• 7–8: Well-constructed, thorough, and accurate; legal references are apt and sufficiently explained.\\
• 9–10: Exceptionally accurate, comprehensive, and displays strong understanding of legal principles\\
\\
\#\# \textbf{Output Format}\\
Return your evaluation results strictly in the following JSON format, providing only the scores:\\
\{ \\
  "Correctness": score (1–10),\\
  "Completeness": score (1–10),\\
  "Clarity": score (1–10),\\
  "Consistency with Provided Passage": score (1–10),\\
\}\\
\end{minipage}
}
}
\caption{Prompt template for \textbf{QA} task evaluation.}
\label{tab:prompt_qa_gen}
\end{table}
\begin{table}[h]
\centering
\scalebox{1}{
\small
\fbox{
\begin{minipage}{\dimexpr\linewidth-5\fboxsep-2\fboxrule\relax}
You are an experienced legal expert evaluating an MRC (Machine Reading Comprehension) response. The system is given a question and a legal passage, and it generates an answer. Please compare the AI’s answer to the reference “gold” answer.\\
\\
\#\# \textbf{Provided Materials}\\
1. Query: The user's question and legal passage.\\
2. AI Response: The AI assistant's response to be evaluated.\\
3. Gold Answer: The ideal response to the `Query`.\\
\\
\#\# \textbf{Evaluation Criteria} (each scored 1–10):\\
1. Correctness: Does the `AI Response` accurately match the `Gold Answer`, especially regarding legal facts and conclusions?\\
2. Completeness: Does the `AI Response` address all parts of the `Query`, covering key details?\\
3. Clarity: Is the `AI Response` written in a clear, unambiguous manner?\\
4. Consistency with Provided Passage: Does the `AI Response` rely on or align with the source passage in the `Query`, avoiding hallucinations?\\
\\
\#\# \textbf{Scoring Guide}:\\
• 1–2: Major factual errors, missing essential information, or irrelevance.\\
• 3–4: Partially correct but lacking thoroughness or clarity.\\
• 5–6: Reasonably accurate, with only minor gaps or ambiguities.\\
• 7–8: Good alignment with the question and passage, minimal issues.\\
• 9–10: Fully accurate, comprehensive, and seamlessly consistent with the gold answer.\\
\\
\#\# \textbf{Output Format}\\
Return your evaluation results strictly in the following JSON format, providing only the scores:\\
\{ \\
  "Correctness": score (1–10),\\
  "Completeness": score (1–10),\\
  "Clarity": score (1–10),\\
  "Consistency with Provided Passage": score (1–10),\\
\}\\
\end{minipage}
}
}
\caption{Prompt template for \textbf{MRC} task evaluation.}
\label{tab:prompt_mrc}
\end{table}

\end{document}